\documentclass{article}
\usepackage{ijcai19}

\usepackage{times}
\usepackage{soul}
\usepackage{url}
\usepackage[hidelinks]{hyperref}
\usepackage[utf8]{inputenc}
\usepackage[small]{caption}
\usepackage{graphicx}
\usepackage{amsmath}
\usepackage{booktabs}
\usepackage[ruled,linesnumbered]{algorithm2e}
\usepackage{latexsym} 
\usepackage{enumitem}
\usepackage{subcaption} 

\usepackage{pgfplots}
\pgfplotsset{compat=newest}
\definecolor{crimson}{rgb}{0.86, 0.08, 0.24}
\definecolor{carrotorange}{rgb}{0.93, 0.57, 0.13}
\definecolor{goldenyellow}{rgb}{1.0, 0.87, 0.0}
\definecolor{darkspringgreen}{rgb}{0.09, 0.45, 0.27}
\definecolor{denim}{rgb}{0.08, 0.38, 0.74}
\definecolor{darkcerulean}{rgb}{0.03, 0.27, 0.49}
\definecolor{dandelion}{rgb}{0.94, 0.88, 0.19}
\usepackage{graphicx}
\usepackage{tikz}

\usepackage{amsmath}
\usepackage{xfrac}    
\usepackage{amsmath}  
\usepackage{nicefrac} 
\newcommand{\rom}[1]{%
  \textup{\uppercase\expandafter{\romannumeral#1}}%
}

\usepackage{booktabs}       
\usepackage{multirow}
\usepackage{longtable}
\usepackage{siunitx}

\title{Multiple Policy Value Monte Carlo Tree Search}

\author{
Li-Cheng Lan$^{1,2}$\and
Wei Li$^{1}$\and
Ting-Han Wei$^{1,2}$\And
I-Chen Wu$^{1,2}$
\affiliations
$^1$National Chiao Tung University, Taiwan \\
$^2$Pervasive Artificial Intelligence Research (PAIR) Labs, Taiwan
\emails
\{sb710031, fm.bigballon, tinghan.wei\}@gmail.com, icwu@csie.nctu.edu.tw
}

\begin{document}

\maketitle

\begin{abstract}
 
Many of the strongest game playing programs use a combination of Monte Carlo tree search (MCTS) and deep neural networks (DNN), where the DNNs are used as policy or value evaluators. Given a limited budget, such as online playing or during the self-play phase of AlphaZero (AZ) training, a balance needs to be reached between accurate state estimation and more MCTS simulations, both of which are critical for a strong game playing agent. Typically, larger DNNs are better at generalization and accurate evaluation, while smaller DNNs are less costly, and therefore can lead to more MCTS simulations and bigger search trees with the same budget. This paper introduces a new method called the multiple policy value MCTS (MPV-MCTS), which combines multiple policy value neural networks (PV-NNs) of various sizes to retain advantages of each network, where two PV-NNs $f_{S}$ and $f_{L}$ are used in this paper. We show through experiments on the game NoGo that a combined $f_{S}$ and $f_{L}$ MPV-MCTS outperforms single PV-NN with policy value MCTS, called PV-MCTS. Additionally, MPV-MCTS also outperforms PV-MCTS for AZ training.

\end{abstract}

\section{Introduction}
\label{Intro}

Many of the state-of-the-art game playing programs in games such as Go, chess, shogi, and Hex used a combination of Monte Carlo tree search (MCTS) and deep neural networks (DNN) \cite{silver2016mastering,silver2018general,gao2018three}. MCTS is a heuristic best-first search algorithm and had been a major breakthrough for many games since 2006 \cite{kocsis2006bandit,coulom2006efficient,browne2012survey}, especially for Go \cite{gelly2007combining,enzenberger2010fuego}. Starting in 2015, DNNs were investigated for move prediction \cite{clark2015training,tian2015better}. For a problem with large state spaces like Go, the power of DNNs to generalize for previously unseen states is critical. In addition, DNNs were found to be more accurate at evaluating positions than Monte Carlo rollout \cite{silver2016mastering}. MCTS, however, remains a critical component of strong game playing programs due to its ability to search ahead while balancing exploration and exploitation.

The choice of network size in an MCTS and DNN combined algorithm is a non-trivial decision. Common current practices tend to settle on a network size based on empirical experience; for the example of Go, most teams tend to settle on 256 filters with varying number of layers \cite{tian2019elfopengo,Leela-zero}, following the first example by AlphaGo \cite{silver2016mastering}. Empirical data shows that overall, the larger the network, the stronger the program tends to be under the same number of MCTS simulations \cite{Leela-zero}. Similarly, outside the context of Go, there is abundant evidence showing that in general, the larger the network, the better it will be at generalization, and the increased capacity of the network also leads to better learned representations \cite{hinton2015distilling,he2016deep}.

However, when creating a program that combines MCTS and DNN, given the same amount of computing resources, it is not a simple decision of training using the largest allowable network, since the number of MCTS simulations depends on the size of the network. A smaller network can be much faster and therefore search more states; given the asymptotic convergence guarantee of MCTS, it may be more favorable to spend the finite budget on performing more simulations, rather than using a larger DNN for a more accurate evaluation. When considering training following the AZ paradigm, this problem is critical, since millions of self-play records need to be generated.

To solve this dilemma, we consider taking advantage of both large and small networks by combining them together with MCTS, which we call multiple policy value MCTS (MPV-MCTS). Two pre-trained DNNs are used in this paper, typically of significantly different network sizes (i.e., consisting of different numbers of filters and layers). MPV-MCTS is a general method that does not depend on how the two DNNs are trained. In the method, each network grows its own best-first search tree. The two trees share the same action value, so intuitively, the small net helps avoid blind spots via its lookahead from a higher simulation count, while the large net provides more accurate values and policies.

We use a simplified variant of Go called NoGo \cite{nogo} to demonstrate the idea. Experiments show that by combining supervised learning networks of different sizes, MPV-MCTS can reach an average of 418 Elo rating against the baseline (HaHaNoGo), whereas a small network and a large network alone can only reach 277 and 309 Elo ratings, respectively. Compared with intermediate-sized single networks, MPV-MCTS is also stronger against the common baseline. MPV-MCTS can also improve the playing strength of separately trained AZ networks. Lastly, using equivalent training budgets, MPV-MCTS can accelerate AZ training by at least a factor of 2. Matchups between the MPV-MCTS method and those without achieves win rates of 56.6\% and 51.2\% for 2 times and 2.5 times the training budget, respectively. With the same training budget, AZ training with MPV-MCTS can be up to 252 Elo ratings stronger than those trained without.

We list two major contributions of this paper as follows:
\begin{enumerate}
    \item The MPV-MCTS search tree generated with two differently sized DNNs $f_S$ and $f_L$ is stronger in playing strength than either net alone, given the same amount of computing resource usage.
    \item With MPV-MCTS, AZ training can be performed more efficiently.
\end{enumerate}

\section{Background}
\label{Background}

\subsection{Monte Carlo Tree Search}
\label{MCTS}
MCTS is a best-first tree search algorithm, typically consisting of the iterative process of: 1) traversing the search tree according to a specified selection algorithm to \emph{select} a yet-to-be-evaluated leaf state
2) \emph{evaluating} the leaf, and 3) \emph{updating} the tree correspondingly with the evaluation result. 
Instead of following the minimax paradigm, MCTS averages the evaluation results in the subtree rooted at state $s$ to decide on the state value $V(s)$.
Monte Carlo sampling (often referred to as rollout) is a standard evaluation method; modern programs may use DNNs as state value approximators instead of rollout, or combine the two evaluation methods in some way. Most selection algorithms are designed to minimize the expected regret of sampling the state $s$ by balancing exploration and exploitation. Examples of selection algorithms include UCB1 \cite{kocsis2006bandit} and PUCT \cite{rosin2011multi}. Both of these algorithms follow the general form:

\begin{equation}
    a_t=\mathop{\arg\max}_{a} (Q(s,a)+u(s,a)),
\end{equation}

where $Q(s,a)$ is the state action value of taking action $a$ at $s$, and $u(s,a)$ is a bonus value to regulate exploration. For example, the bonus value of PUCT (used by AlphaGo) is:
\begin{equation}
\label{equ:PUCT}
    u_{\text{PUCT}}(s,a)\propto P(s,a)\times\frac {\sqrt{N(s)}}{N(s,a)},
\end{equation}
where $N(s)$ and $N(s,a)$ are the simulation counts of $s$ and taking the action $a$ at $s$ respectively; and
$P(s,a)$ is the probability of $a$ being the best action of $s$, which is referred to as the \emph{prior probability}.

\subsection{Policy Value MCTS}
\label{PV-MCTS}
Policy Value MCTS (PV-MCTS) uses DNNs to provide a policy $p(a|s)$ and a value $v(s)$ for any given state $s$. The policy $p$ is used for PUCT as $P(s,a)$ in Equation \ref{equ:PUCT} during the selection phase, while $v(s)$ is used as the evaluation result to update the state value $V$ of ancestor states of $s$. The particular implementation of PV-MCTS as used by AlphaGo consists of two separate networks, the policy networks (be used when a node become a branch node of MCTS) and value networks (be used when a node be selected). AlphaGo Zero combines the two as a policy value network (PV-NN) that outputs the prior probabilities and state value simultaneously.

Data from Silver et al. \shortcite{silver2017mastering} hinted that the advantage of search cannot be easily replaced by spending more effort on training better policy networks. More specifically, their experiments show that the output policy of PV-MCTS is about 2000 Elo ratings \cite{elo1978rating} stronger than simply using the policy output of the same PV-NN without search. This indicates that even with a strong DNN policy function, performance can be further improved significantly through search. 

\subsection{Combining Multiple Strategies in MCTS}
\label{APV-MCTS}
There are several methods that have been proposed to combine multiple strategies during MCTS action value evaluation, such as Rapid Action Value Evaluation (RAVE) \cite{gelly2011monte}, implicit minimax backups \cite{lanctot2014monte}, and asynchronous policy value MCTS (APV-MCTS) \cite{silver2016mastering}. Both RAVE and implicit minimax backup combine MCTS evaluation (rollout) with another heuristic; for the scope of this paper, we omit the details. Of these three algorithms, MPV-MCTS is most related to APV-MCTS. APV-MCTS is designed to combine the value network and rollout as the MCTS evaluation. The combined strength of the value network and rollout can achieve more than 95\% winning rate against using either one alone \cite{silver2016mastering}, demonstrating that there is potential for better performance by combining different strategies.



\section{Method}
\label{Method}



\subsection{Multiple Policy Value MCTS}
\label{MPV-MCTS}
In this subsection, we focus on explaining the scenario where the overall system consists of two PV-NNs $f_S$ and $f_L$ (small and large networks, respectively). Let $b_S$ ($b_L$) be an assigned number of simulations, or budget, for which our method uses the network $f_S$ ($f_L$) to grow its own search tree $T_S$ ($T_L$). Our problem is then: Given $s$, $f_S$, $f_L$, $b_S$, and $b_L$, find a stronger policy:
$$\pi(s,(f_S,b_S),(f_L,b_L)),$$ such that $b_S \geq b_L$.


When a state is in both search trees, the two networks collaborate by sharing the same state value $V(s)$ and the same prior probability $P(s,a)$.
For the scope of this paper, we use the following method, similar to APV-MCTS:
\begin{align}
    V(s) &= \alpha V_S(s)+(1-\alpha) V_L(s), \text{ and} \\
    P(s,a) &= \beta p_S(a|s)+(1-\beta) p_L(a|s),
\end{align}
where $\alpha, \beta \in [0,1]$ are weight coefficients. Note that $\alpha$ and $\beta$ can be set according to the accuracy of the values and prior probabilities. For example, the more accurate $V_L$ and $p_L$ are, the smaller $\alpha$, $\beta$ should be. In the experiments, we set $\alpha=0.5$, $\beta=0$, following settings by APV-MCTS.

For each simulation during the search, we first choose either $f_S$ or $f_L$, say $f_S$, then select a leaf state of $T_S$ to evaluate, and then use the results to update the search tree. It is up to the user to design how the two networks take turns, as long as the budgets $b_S$ and $b_L$ are satisfied. We suggest several ways to take turn in section \ref{Discussion}. In our experiments, for a given set of $b_S$ and $b_L$, we simply randomly select $b_S$ numbers between 1 and $b_S+b_L$, and perform a small net simulation for those iterations; for all other iterations, we use the larger net. This corresponds to line 1 in Algorithm \ref{alg:MPV-MCTS}.

We now consider how $f_S$ and $f_L$ contribute to the overall MPV-MCTS method, starting with $f_S$. Conceptually, the goal is to allow $T_S$ to provide the overall MPV-MCTS with the benefits of lookahead search, where the tree balances between exploration and exploitation as it grows. This is the role of $f_S$ because it is the faster of the two networks, and can therefore perform more simulations with the same amount of resource usage as $f_L$. This is also the reason why we assign $b_S$ to be larger than $b_L$. During MPV-MCTS, for each simulation using $f_S$, a leaf state is selected following the PUCT algorithm using the simulation count of $f_S$ as $N$. This corresponds to lines 4-6 in Algorithm \ref{alg:MPV-MCTS}.

Now we consider the role of $f_L$. For every simulation of $f_L$, we wish to identify and simulate the most critical states. While there are many ways to do so, we simply assume for simplicity that in the larger tree the yet-to-be-evaluated leaf with the higher visit count in $T_S$ should be more important, and thus the leaf with the highest visit count in $T_S$ is selected in each simulation. This corresponds to line 8 in Algorithm \ref{alg:MPV-MCTS}. There is a very rare special case, wherein the selected leaf may not yet be visited by $f_S$. In this case (line 9), we reselect a unevaluated leaf state (line 10) using PUCT for $f_L$ instead (following Equation \ref{equ:PUCT}, but with $N_L(s)$ and $N_L(s,a)$).

\begin{algorithm}
\KwIn{state $s$, networks $f_S$ and $f_L$, budgets $b_S$ and $b_L$}
$list = \text{RandomlySelect}(b_S, b_S+b_L)$ \;
 \For{$i\gets1$ \KwTo $b_S+b_L$ }{
  \uIf{ $i$ \text{in} $list$ }{
  $s_{leaf}  = \text{SelectUnevaluatedLeafStateByPUCT}(T_S)$ \;
  $(p,v) = f_S(s_{leaf})$ \;
  $\text{Update}(T_S,s_{leaf} ,(p,v))$ \;
  }
  \Else{
  $s_{leaf} = \text{SelectUnevaluatedLeafStateByPriority}(T_L)$ \;
  \uIf{$N_S(s_{leaf}) = 0$}{
   $s_{leaf}  = \text{SelectUnevaluatedLeafStateByPUCT}(T_L)$ \;
  }
  $(p,v) =f_L(s_{leaf})$\;
  $\text{Update}(T_L,s_{leaf},(p,v))$\;
  }
 }
 \caption{MPV-MCTS Algorithm}
 \label{alg:MPV-MCTS}
\end{algorithm}

\subsection{Training AlphaZero with MPV-MCTS}
\label{Training AlphaZero with MPV-MCTS}
We first briefly review the AZ training method, in which an agent improves by playing against itself repeatedly. The agent generates its playing policy by using a PV-NN in a PV-MCTS algorithm. The weights in the PV-NN are first initialized randomly. We now describe the training process by listing three major components.
\begin{itemize}
    \item {By using the PV-NN in a PV-MCTS algorithm, \emph{self-play workers} generate game records by letting two same instances of itself play against each other. At each turn in a game, a worker computes PV-MCTS and follows the playing policy as follows. The probability of $a$ being played at $s$ is $\pi_a \propto  N(s,a)^{1/\tau}$, where $N(s,a)$ is the action's simulation count and $\tau$ is a temperature parameter. When a game ends, the self-play record is saved to the replay buffer.}
    \item {The \emph{replay buffer} is a fixed-size queue containing a collection of self-play game records. Each record $(s, \pi, z)$ includes a state $s$ (of which there will be many in a single game), the playing policy $\pi$ used to select the action at state $s$, and the outcome $z$ of the game.}
    \item {The \emph{training worker} continually samples records from the replay buffer and trains the current PV-NN.}
\end{itemize}

Following the above AZ training and from the intuition that larger DNNs tend to learn better, it is reasonable to expect the trained agent using $f_L$ should outperform that with $f_S$. That said, if our hypothesis for MPV-MCTS holds, it is also possible that replacing PV-MCTS with MPV-MCTS in AZ training will lead to better performance.

\begin{figure}[h]
\centering
\includegraphics[width=7cm]{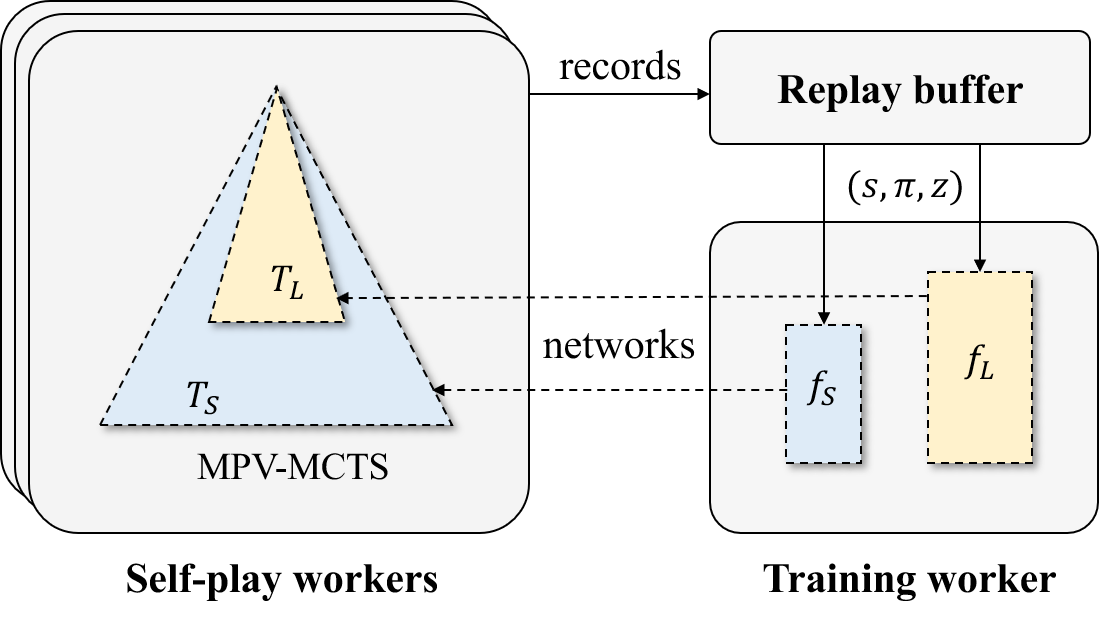}
\caption{ Workflow of training AZ with MPV-MCTS.  }
\label{fig:workflow}
\end{figure}

Figure \ref{fig:workflow} shows the work-flow of how MPV-MCTS can be applied to the AZ training algorithm. We start from two PV-NNs of different sizes with random weights. The replay buffer does not need to be modified. The self-play workers use the latest PV-NNs of both sizes with MPV-MCTS to generate the self-play records. After performing MPV-MCTS, we simply use the simulation count of the small net tree $T_S$ as our playing policy $\pi$ to select actions. When a game ends, we store the records to the replay buffer. The training worker repeatedly use records sampled from the replay buffer to train both $f_S$ and $f_L$ using the same loss function as AlphaGo Zero \cite{silver2017mastering}.

\section{Experiments}
\label{Experiments}

In this section, we empirically demonstrate our method on NoGo, a two-player game on a $9 \times 9$ Go board. NoGo was selected as the feature game of the 2011 Combinatorial Games Workshop at BIRS, and subsequently became a tournament item in the Computer Olympiad \cite{nogo}. The game has the same rules as Go, except where moves that lead to capturing and suicide are not allowed. The player who cannot make any moves during her turn loses. 
NoGo is chosen for our experiments since the complexity of the game is relatively low when compared with Go, while maintaining many similar characteristics \cite{chou2011revisiting}. 


\subsection{Experiment Settings}
In our experiments, the state-of-the-art NoGo program HaHaNoGo \cite{HaHaNoGo} served as the benchmark. HaHaNoGo is a MCTS-based program that defeated the reigning champion HappyNoGo (which placed first in the 2013 and 2015 Computer Olympiad) in a competition held in TAAI 2016. In all experiments, we used HaHaNoGo with 100,000 simulations as the baseline, and each program played 1,000 games (the standard deviation of the win rate is about 1\%) against the baseline to obtain a win rate. For better visualization and analysis, we use the Elo rating \cite{elo1978rating} instead of win rate to judge the playing strength of AI by setting the Elo rating of HaHaNoGo to $0$. In this paper, all experiments are performed on eight Intel Xeon(R) Gold 6154 CPUs and 64 Nvidia Tesla V100 GPUs.

The architecture of our PV-NNs is the same as that used in AlphaGo Zero \cite{silver2017mastering}, except for filter sizes, residual blocks \cite{he2016deep} and inputs. Let $f(x,y)$ denote a PV-NN with $x$ filters and $y$ residual blocks. The input to the network is a $9 \times 9 \times 4$ image stack comprised of four binary feature planes for the board information (player's stones, opponent's s, player's legal moves, opponent's moves) representing game states. 

For fairness of comparison, let one unit of normalized budget indicate the amount of computing resources consumed for one single forward pass on network $f_{128,10}$ on the environment of the above setting. Note that one forward pass on $f(a\times x,b\times y)$ ideally runs about $a^2\times b$ times slower than $f(x,y)$. Thus, for computing resource analysis, one single forward pass on $f_{64,5}$ is said to consume $1/8$ normalized budget (ignoring the cost of selection and update). Thus, given the same amount of normalized budget $B$, $B$ forward passes can be performed on $f_{128,10}$, and $8B$ on $f_{64,5}$.

\subsection{Evaluating MPV-MCTS}
\label{sec:evaluating MPV-MCTS}
\paragraph{Combining Supervised Learning Networks.}
First, we investigate using PV-NNs that were trained with supervised learning in MPV-MCTS. We trained both $f_{64,5}$ and $f_{128,10}$ from a dataset of 200,000 games (about $10^7$ positions) generated by HaHaNoGo with 50,000 simulations for each move via self-play. We will present the performance of MPV-MCTS with different budget allocation schemes, compared to single networks using PV-MCTS. 




With a normalized budget of $B$, we combine $f_{64,5}$ and $f_{128,10}$ with the MPV-MCTS algorithm by different allocation schemes according to a budget ratio $r \in [0,1]$, where the normalized budget is $rB$ for $f_{128,10}$ and $(1-r)B$ for $f_{64,5}$. 

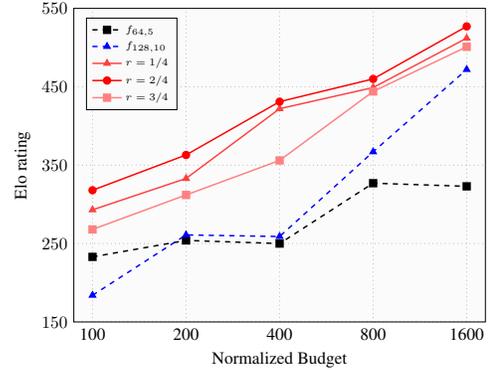
\begin{figure}[htb]

\centering
\begin{tikzpicture}[scale=0.65]
\begin{axis}[
    width=10cm,
    height=8cm,
    mark options={solid},
    xlabel={Normalized Budget},
    ylabel={Elo rating},
    symbolic x coords={100, 200, 400, 800, 1600},
    xmin=100, xmax=1600,
    ymin=150, ymax=550,
    enlarge x limits=0.05,
    xtick={100,200, 400, 800, 1600},
    ytick={150, 250, 350, 450, 550},
    ymajorgrids=true,
    tick style={draw=none},
    grid,
    grid style={dotted,gray!70},
    legend style={font=\tiny,legend pos=north west,cells={anchor=west},fill=gray!4},
    axis background/.style={fill=gray!3}
]
 
 \addplot [
    color=black,
    line width=0.3mm,
    mark=square*,
    dashed
    ]
    coordinates {
    (100,233)(200,254)(400,250)(800,327)(1600,323)
    };
    \addlegendentry{$f_{64,5}$}
 \addplot [
    color=blue,
    line width=0.3mm,
    mark=triangle*,
    dashed
    ]
    coordinates {
    (100,184)(200,261)(400,259)(800,367)(1600,472)

    };
    \addlegendentry{$f_{128,10}$}

 \addplot [    
    color=red!75,
    mark=triangle*,
    line width=0.3mm,
    ]
    coordinates {
    (100,293)(200,333)(400,422)(800,449)(1600,512)
    };
    \addlegendentry{$r=1/4$}

  \addplot [
    color=red,
    line width=0.3mm,
    mark=*,
    ]
    coordinates {
    (100,318)(200,363)(400,431)(800,460)(1600,527)
    };
    \addlegendentry{$r=2/4$}
  
 \addplot [
    color=red!50,
    line width=0.3mm,
    mark=square*,
    ]
    coordinates {
    (100,268)(200,312)(400,356)(800,444)(1600,501)

    };
    \addlegendentry{$r=3/4$}
    
\end{axis}
\end{tikzpicture}
\caption{The Elo ratings for different kinds of (normalized) budget allocation. For a given budget $B$, $rB$ is allocated to $f_{128,10}$, while the remaining $(1-r)B$ is allocated to network $f_{64,5}$. When $r=1$, it is equivalent to PV-MCTS with $f_{128,10}$ with simulation count $B$; when  $r=0$, it is equivalent to $f_{64,5}$ with simulation count $8B$.}
\label{fig:sl-haha-allocation}

\end{figure}

Figure \ref{fig:sl-haha-allocation} shows the result of different resource allocation schemes, showing that combining two PV-NNs leads to better performance. The strongest version in these experiments is $r=2/4$, which achieved $527$ Elo rating (about $95.40\%$ win rate against the baseline) with a normalized budget of 1600. While the version that only uses either the small or the large network can only achieve $323$ and $472$ Elo rating at best, respectively. With the same normalized budget, both PV-NNs $f_{64,5}$ and $f_{128,10}$ alone are weaker than all three allocation ratios of MPV-MCTS. In the rest of the experiment, we set $r=2/4$. 

We also trained intermediate-sized PV-NNs, $f_{128,5}$, $f_{90,10}$, $f_{90,5}$, and $f_{64,10}$, which are expected to be more accurate than smaller PV-NNs (e.g., $f_{64,5}$) and faster than larger PV-NNs (e.g., $f_{128,10}$). We compared these mid-sized PV-NNs using PV-MCTS with the best performing set in Figure \ref{fig:sl-haha-allocation} (i.e., MPV-MCTS with $f_{64,5}+f_{128,10}, r=2/4$) with the same normalized budget.  Figure \ref{fig: mid vs combine} shows that MPV-MCTS outperforms all of the mid-sized PV-NNs with PV-MCTS. 

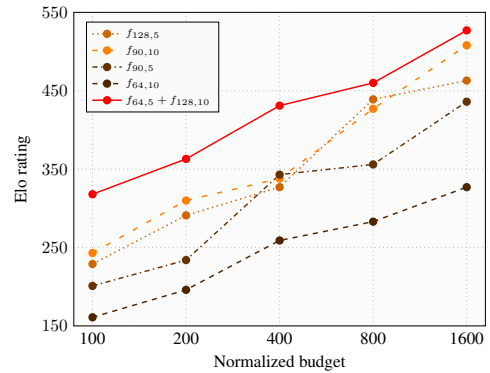
\begin{figure}[htb]
\centering
\begin{tikzpicture}[scale=0.65]
\begin{axis}[
    width=10cm,
    height=8cm,
    mark options={solid},
    xlabel={Normalized budget},
    ylabel={Elo rating},
    symbolic x coords={100, 200, 400, 800, 1600},
    xmin=100, xmax=1600,
    ymin=150, ymax=550,
    enlarge x limits=0.05,
    xtick={100,200, 400, 800, 1600},
    ytick={150, 250, 350, 450, 550},
    ymajorgrids=true,
    tick style={draw=none},
    grid,
    grid style={dotted,gray!70},
    legend style={font=\tiny,legend pos=north west,cells={anchor=west},fill=gray!4},
    axis background/.style={fill=gray!3}
]
 
  \addplot [    
    color=orange!80!black,
    line width=0.3mm,
    mark=*,
    dotted
    ]
    coordinates {
    (100,229)(200,291)(400,327)(800,439)(1600,463)
    };
    \addlegendentry{$f_{128,5}$}

 \addplot [
    color=orange,
    line width=0.3mm,
    mark=*,
    loosely dashed
    ]
    coordinates {
    (100,243)(200,310)(400,338)(800,427)(1600,508)

    };
    \addlegendentry{$f_{90,10}$}

 \addplot [
    color=orange!40!black,
    line width=0.3mm,
    mark=*,
    dashdotted
    ]
    coordinates {
    (100,201)(200,234)(400,343)(800,356)(1600,436)
    };
    \addlegendentry{$f_{90,5}$}

 \addplot [
    color=orange!30!black,
    line width=0.3mm,
    mark=*,
    dashed
    ]
    coordinates {
    (100,161)(200,196)(400,259)(800,283)(1600,327)
    };
    \addlegendentry{$f_{64,10}$}
    
  \addplot [
    color=red,
    line width=0.3mm,
    mark=*,
    ]
    coordinates {
    (100,318)(200,363)(400,431)(800,460)(1600,527)
    };
    \addlegendentry{$f_{64,5}+f_{128,10}$}
    
\end{axis}
\end{tikzpicture}
\caption{The Elo ratings of intermediate-sized networks with PV-MCTS vs. $f_{64,5}+f_{128,10}$ with MPV-MCTS.}\label{fig: mid vs combine}
\end{figure}
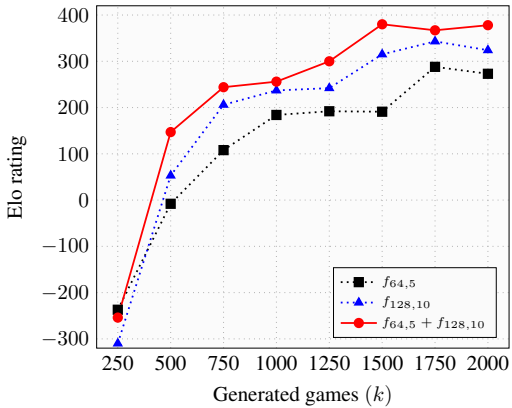
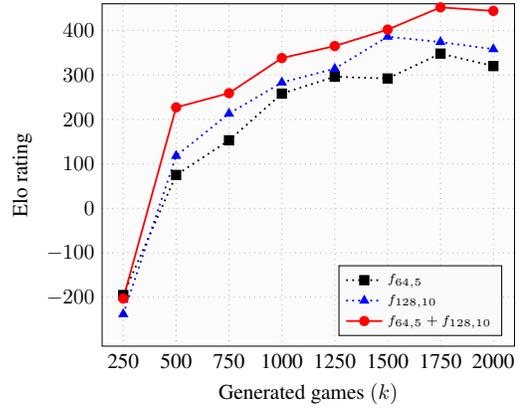
\begin{figure*} [hbt]
    \centering

\begin{subfigure}[hbt]{0.48\textwidth}
        \centering
        \begin{tikzpicture}[scale=0.8]
        \begin{axis}[
            mark options={solid},
            xlabel={Generated games $(k)$},
            ylabel={Elo rating},
            xmin=150, xmax=2100,
            ymin=-320, ymax=420,
            xticklabels={250,500,750,1000,1250,1500,1750,2000},
            ytick={-300,-200,-100,0,100,200,300,400},
            xtick=data,
            ymajorgrids=true,
            tick style={draw=none},
            grid,
            grid style={dotted,gray!70},
            legend style={font=\tiny,legend pos=south east,cells={anchor=west},fill=gray!4},
            axis background/.style={fill=gray!3}
        ]
         \addplot [
            color=black,
            mark=square*,
            line width=0.3mm,
            dotted
            ]
            coordinates {
            (250,-237)(500,-8)(750,108)(1000,184)(1250,192)(1500,191)(1750,288)(2000,273)
            };
            \addlegendentry{$f_{64,5}$}
         \addplot [
            color=blue,
            mark=triangle*,
            line width=0.3mm,
            dotted
            ]
            coordinates {
            (250,-310)(500,53)(750,206)(1000,237)(1250,242)(1500,315)(1750,343)(2000,324)
            };
            \addlegendentry{$f_{128,10}$}
         \addplot [
            color=red,
            mark=*,
            line width=0.3mm,
            ]
            coordinates {
            (250,-254)(500,147)(750,244)(1000,256)(1250,300)(1500,380)(1750,367)(2000,378)
            };
            \addlegendentry{$f_{64,5}+f_{128,10}$}

        \end{axis}
        \end{tikzpicture}
        \caption{Testing with a normalized budget of $200$.}\label{difVW1x}
        \label{fig:1x simulation}
    \end{subfigure}
    \hfill
\begin{subfigure}[hbt]{0.48\textwidth}
        \centering
        \begin{tikzpicture}[scale=0.8]
        \begin{axis}[
            mark options={solid},
            xlabel={Generated games $(k)$},
            ylabel={Elo rating},
            xmin=150, xmax=2100,
            ymin=-310, ymax=460,
            xticklabels={250,500,750,1000,1250,1500,1750,2000},
            ytick={-200,-100,0,100,200,300,400},
            xtick=data,
            ymajorgrids=true,
            tick style={draw=none},
            grid,
            grid style={dotted,gray!70},
            legend style={font=\tiny,legend pos=south east,cells={anchor=west},fill=gray!4},
            axis background/.style={fill=gray!3}
        ]
         \addplot [
            color=black,
            mark=square*,
            line width=0.3mm,
            dotted
            ]
            coordinates {
            (250,-195)(500,75)(750,153)(1000,258)(1250,296)(1500,292)(1750,348)(2000,320)
            };
            \addlegendentry{$f_{64,5}$}
         \addplot [
            color=blue,
            mark=triangle*,
            line width=0.3mm,
            dotted
            ]
            coordinates {
            (250,-238)(500,118)(750,213)(1000,283)(1250,314)(1500,386)(1750,374)(2000,358)
            };
            \addlegendentry{$f_{128,10}$}
         \addplot [
            color=red,
            mark=*,
            line width=0.3mm,
            ]
            coordinates {
            (250,-203)(500,227)(750,259)(1000,338)(1250,365)(1500,402)(1750,452)(2000,444)
            };
            \addlegendentry{$f_{64,5}+f_{128,10}$}

        \end{axis}
        \end{tikzpicture}
        \caption{Testing with a normalized budget of $800$.}\label{difVW4x}
        \label{fig:4x simulation}
    \end{subfigure}
    \caption{Combining PV-NNs trained by AZ with MPV-MCTS, where each red dot is the combination of the blue and black dots in the same column.}
    \label{fig: MPV-MCTS use PV-NN trained by AlphaZero}
\end{figure*}

\paragraph{Combining AlphaZero Trained Networks.}
\label{sec:az training}
Since MPV-MCTS is comprised of pre-trained PV-NNs, we now investigate using various AZ trained PV-NNs in pairs. We trained both $f_{64,5}$ and $f_{128,10}$ using the following settings: 
\begin{itemize}
    \item self-play simulation: $800$
    \item PUCT constant $c_{\text{PUCT}}$: $1.5$
    \item replay buffer size: $100,000$
\end{itemize}
We used 60 GPUs as self-player workers and 4 GPUs as a training worker. Each network was trained with about 500,000 steps, and 2,000,000 generated self-play games.

After training, $f_{64,5}$ and $f_{128,10}$ will each have a progression of PV-NNs as the accumulated number of self-play games approaches 2,000,000. We select eight pairs of large and small PV-NNs, along the following total number of accumulated self-play games: 250k, 500k, ... , 2000k. 
That is, in Figure \ref{fig: MPV-MCTS use PV-NN trained by AlphaZero}, the red data points refer to the MPV-MCTS consisting of $f_{64,5}$ and $f_{128,10}$, each trained following AZ with 250k accumuluated self-play game records. Figure \ref{fig: MPV-MCTS use PV-NN trained by AlphaZero} also presents the strength of each pair of large and small PV-NNs using the same normalized budget for testing. For Figure \ref{fig:1x simulation}, a normalized budget of 200 was used, while for Figure \ref{fig:4x simulation}, a normalized budget of 800 was used. Both results show that MPV-MCTS outperforms both the large and small nets, at all stages throughout the AZ training process. This also shows the robustness of MPV-MCTS. 

\subsection{AlphaZero Training with MPV-MCTS}

In this subsection, we investigate performing AZ training with MPV-MCTS, as illustrated in Figure \ref{fig:workflow}. 
Let $f_{S}$ be $f_{64,5}$, and $f_{L}$ be $f_{128,10}$ for clarity of presentation. Following the workflow in Figure \ref{fig:workflow}, both $f_L$ and $f_S$ are trained together; self-play workers use both PV-NNs to generate self-play games, using the playing policy $\pi(s,(f_S,800)),(f_L,100)$ (as in the terminology in subsection \ref{MPV-MCTS}). 

Next, we follow AZ training for three separate large networks, where the difference is that during self-play, each move is generated with simulation counts of 200, 400, and 800. These three are denoted as $f_{L_{200}}$, $f_{L_{400}}$, and $f_{L_{800}}$, respectively. Since we wish to fix the total training resource usage, we will need to define a \emph{normalized game generation count}, similar to the way we defined a normalized budget. Since the version with 800 simulations per self-play move theoretically spends four times approximately as much resources as the version with 200 simulations per self-play, the former will generate $1/4$ as many self-play game records given the same training budget. For this experiment, we define 1 normalized generated game to use the same amount of training budget as generating 1 self-play game record for $f_{128,10}$ using 200 simulations per move. Note that in the MPV-MCTS case (where $f_S$ and $f_L$ are trained with 800 and 100 simulations per move respectively), generating 1 actual self-play game record is equivalent in terms of cost as 1 normalized generated game.

\begin{figure*} [hbt]
    \centering

\begin{subfigure}[hbt]{0.48\textwidth}
\centering
\begin{tikzpicture}[scale=0.8]
\begin{axis}[
    mark options={solid},
    xlabel={Normalized generated games},
    ylabel={Elo rating},
    xmin=150, xmax=2100,
    ymin=-1250, ymax=450,
    xticklabels={250,500,750,1000,1250,1500,1750,2000},
    ytick={-1200,-800,-400,0,400},
    xtick=data,
    ymajorgrids=true,
    tick style={draw=none},
    grid,
    grid style={dotted,gray!70},
    legend style={font=\tiny,legend pos=south east,cells={anchor=west},fill=gray!4},
    axis background/.style={fill=gray!3}
]

 \addplot [
    color=blue!77,
    mark=square*,
    line width=0.3mm,
    dotted
    ]
    coordinates {
    (250,-1100)(500,-623)(750,-369)(1000,-159)(1250,-119)(1500,-72)(1750,-48)(2000,-16)
    };
    \addlegendentry{$f_{L_{200}}$}
 \addplot [
    color=blue!77,
    mark=*,
    line width=0.3mm,
    dotted
    ]
    coordinates {
    (250,-1079)(500,-494)(750,-251)(1000,-193)(1250,-159)(1500,-149)(1750,-41)(2000,-28)
    };
    \addlegendentry{$f_{L_{400}}$}
 \addplot [
    color=blue!77,
    mark=triangle*,
    line width=0.3mm,
    dotted
    ]
    coordinates {
    (250,-1100)(500,-592)(750,-398)(1000,-310)(1250,-70)(1500,-38)(1750,53)(2000,53)
    };
    \addlegendentry{$f_{L_{800}}$}

 \addplot [
    color=green!70!black,
    mark=*,
    line width=0.3mm,
    ]
    coordinates {
    (250,-305)(500,-12)(750,89)(1000,122)(1250,219)(1500, 277)(1750,272)(2000,279)
    };
    \addlegendentry{$f_{S_{800},L_{100}}$  }  
    
\end{axis}
\end{tikzpicture}
\caption{Testing with a normalized budget of $200$.}
\label{fig: Az training with MPV-MCTS 200}
\end{subfigure}
    \hfill
\begin{subfigure}[hbt]{0.48\textwidth}
\centering
\begin{tikzpicture}[scale=0.8]
\begin{axis}[
    mark options={solid},
    xlabel={Normalized generated games},
    ylabel={Elo rating},
    xmin=150, xmax=2100,
    ymin=-1250, ymax=450,
    xticklabels={250,500,750,1000,1250,1500,1750,2000},
    ytick={-1200,-800,-400,0,400},
    xtick=data,
    ymajorgrids=true,
    tick style={draw=none},
    grid,
    grid style={dotted,gray!70},
    legend style={font=\tiny,legend pos=south east,cells={anchor=west},fill=gray!4},
    axis background/.style={fill=gray!3}
]

 \addplot [
    color=blue!77,
    mark=square*,
    line width=0.3mm,
    dotted
    ]
    coordinates {
    (250,-1100)(500,-449)(750,-207)(1000,-71)(1250,-26)(1500,43)(1750,17)(2000,51)
    };
    \addlegendentry{$f_{L_{200}}$}
 \addplot [
    color=blue!77,
    mark=*,
    line width=0.3mm,
    dotted
    ]
    coordinates {
    (250,-1100)(500,-363)(750,-155)(1000,-147)(1250,-65)(1500,-57)(1750,24)(2000,58)
    };
    \addlegendentry{$f_{L_{400}}$}
 \addplot [
    color=blue!77,
    mark=triangle*,
    line width=0.3mm,
    dotted
    ]
    coordinates {
    (250,-1079)(500,-667)(750,-268)(1000,-238)(1250,-24)(1500,8)(1750,151)(2000,118)
    };
    \addlegendentry{$f_{L_{800}}$}

 \addplot [
    color=green!70!black,
    mark=*,
    line width=0.3mm,
    ]
    coordinates {
    (250,-231)(500,48)(750,108)(1000,202)(1250,308)(1500, 348)(1750,320)(2000,370)
    };
    \addlegendentry{$f_{S_{800},L_{100}}$  }  
    
\end{axis}
\end{tikzpicture}
\caption{Testing with a normalized budget of $800$.}
\label{fig: Az training with MPV-MCTS 800}
\end{subfigure}
    \caption{AZ training with MPV-MCTS. For a fair comparison, the Elo ratings were obtained by performing PV-MCTS using equally sized large networks $f_L$ for all cases. That is, $f_S$ was only used for the green dataset during training, and was omitted for testing.}
    \label{fig: Az training with MPV-MCTS}
\end{figure*}
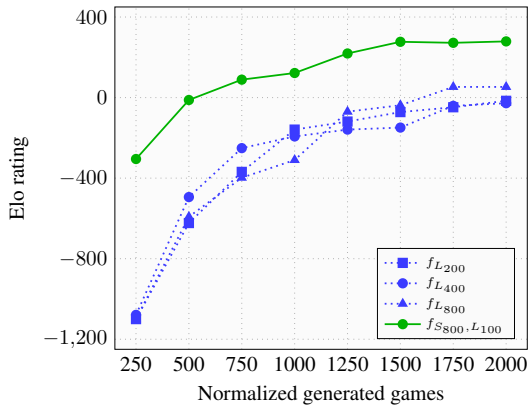
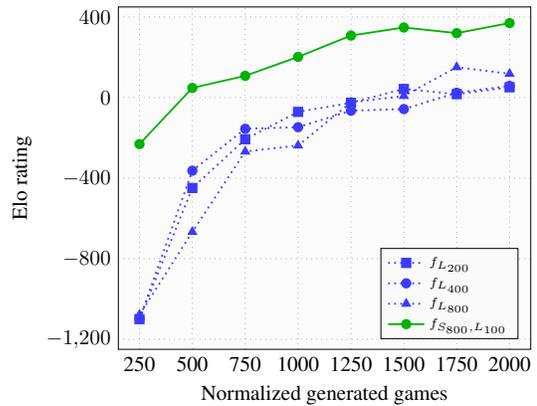

Figure \ref{fig: Az training with MPV-MCTS} presents the results of training AZ with MPV-MCTS compared with the three other large nets. Since our main goal is to demonstrate the benefit of using MPV-MCTS for training, the testing conditions should be as equal across all sets versions as possible. The Elo ratings were obtained by playing against the same baseline (HahaNoGo with 100,000 simulations per move), where each agent with the trained large net uses 200 (Figure \ref{fig: Az training with MPV-MCTS 200}) or 800 (Figure \ref{fig: Az training with MPV-MCTS 800})  simulations per move for testing.
For the MPV-MCTS case, we use only $f_L$ during testing for comparison; $f_S$ was omitted from testing so that the comparison is strictly between large PV-NNs.

The results show that our method outperforms the best large net following AZ training with PV-MCTS (i.e., $f_{L_{800}}$) by 226 and 252 Elo ratings (for 200 and 800 testing simulations) with a total of 2000k normalized generated self-play games. Our interpretation is that this is because with a fixed training budget, the trade-off is again between more simulations and more accurate networks. In the case for $f_{L_{200}}$, the smaller simulation count leads to poor playing policies, consequently making it difficult to improve the network. On the other hand, for $f_{L_{800}}$, the total number of self-play records is $1/4$ that of $f_{L_{200}}$, so the overall learning progress is slow.

The results for $f_{L_{800}}$ in Figure \ref{fig: Az training with MPV-MCTS} corresponds to the blue dataset in Figure \ref{fig: MPV-MCTS use PV-NN trained by AlphaZero}, where the self-play simulation (for training) was also set to be 800. If we look at the MPV-MCTS training case (the green dataset in Figure \ref{fig: Az training with MPV-MCTS}), the result at 2000k normalized generated games performs 279 Elo ratings and 370 Elo ratings stronger than the baseline, for 200 and 800 simulations respectively (Figures \ref{fig: Az training with MPV-MCTS 200} and \ref{fig: Az training with MPV-MCTS 800}). Comparing between the data for MPV-MCTS in Figure \ref{fig: Az training with MPV-MCTS 800} and $f_{128,10}$ in Figure \ref{fig:4x simulation}, the performance for MPV-MCTS still exceeds that of $f_{128,10}$ for 1000k (4000k normalized) and 1250k generated games (5000k normalized). In fact, matches between $f_{S_{800},L_{100}}$ at 2000k normalized generated games and $f_{L_{800}}$ at 1000k generated games yields a win rate of 56.6\%, while playing against $f_{L_{800}}$ at 1250k generated games yields a win rate of 51.2\%. This implies that MPV-MCTS accelerates AZ training by at least factor of 2.

\section{Discussion and Future work}
\label{Discussion}
In this section, we discuss some settings that we only performed partial experiments on, and share preliminary results.

First, instead of two separate trees, we can think of our method as having just one single tree, with the small net’s tree as the “main” tree, because it is usually the larger one. Nonetheless, we describe MPV-MCTS as two trees for generality (thereby allowing the algorithm to work with more than two nets) and simplicity.

Second, in line 8 of Algorithm \ref{alg:MPV-MCTS}, we used the simulation count to select the most important yet-to-be-evaluated states for $f_L$. Other types of priority functions can be used instead when selecting states for $f_L$, such as following its own PUCT, or adding a discounted bonus to states whose parent has higher visit counts. For situations with limited budgets, the former tends to be weaker than our current priority function (but still stronger than a single net), while the latter is complicated but yields no improvement. For multiple simultaneous networks, only the smallest network (with the largest tree) uses PUCT for selection. All other networks would use a user-defined priority function instead.

Third, instead of randomly selecting the order to take turns simulating with $f_S$ and $f_L$, as described in subsection \ref{MPV-MCTS} and line 1 of Algorithm \ref{alg:MPV-MCTS}, we have also tried some other settings. Assume that the budget allocation is $b_S=800, b_L=100$; one alternative method is to force MPV-MCTS to start with the larger net (say, the first 50 simulations all use $f_L$) so that the small net has more information that guides the search when it begins. Results show that this alternative method of taking turns does not yield improvement in testing, but seem to have beneficial effects when used in AZ training (as in subsection \ref{Training AlphaZero with MPV-MCTS}). We also tried round-robin, but no significant difference was observed.

Fourth, our method provides an opportunity for ensemble learning. For example, we tried to increase the coefficient of the mean square error in the small net's loss function, with the aim of improving the accuracy of its value function. Results show that training AZ with MPV-MCTS in this way accelerates the process of training.

Fifth, we trained a mini-net (even weaker than the small net) to replace the small net. The results of 800 mini network simulations + 100 large network simulations were comparable to 800 small network simulations + 100 large network, despite the mini network being obviously worse than the small network. This seems to imply that the MPV-MCTS benefits from the look-ahead search provided by the smaller “partner” more than the quality of the smaller partner.


Finally, we believe that a better way to train the AZ algorithm with MPV-MCTS is to train with the help of several sizes of networks. In this scenario, the largest network is the primary network. The training begins by using the smallest network as the support network, following the training process as the one described in subsection \ref{Training AlphaZero with MPV-MCTS}. As the large network improves and the representation learned by the small network is no longer sufficient to master the training data, we can replace it with a larger support network. This process is iterative; while the primary network is persistent, the support network should increase in capacity whenever it is unable to keep up. The decision of using which support networks and the parameters such as $\alpha$, $\beta$, and simulation count can be controlled by meta-learning. We leave this as an open problem in the future.

\section*{Acknowledgments}

This research is partially supported by the Ministry of Science and Technology (MOST) under Grant Number MOST 107-2634-F-009-011 and MOST 108-2634-F-009-011 through Pervasive Artificial Intelligence Research (PAIR) Labs, Taiwan. The computing resource is partially supported by National Center for High-performance Computing (NCHC).

\bibliographystyle{named}
\bibliography{ijcai19}

\end{document}